\documentclass[11pt,a4paper]{article}
\usepackage[hyperref]{acl2017}
\usepackage{times}
\usepackage{latexsym}
\usepackage{url}
\usepackage{array}
\usepackage{amssymb}
\usepackage{amsmath}
\usepackage{soul}
\usepackage{color}
\usepackage{relsize}
\usepackage{graphicx}
\usepackage{listings}
\usepackage{multirow}
\usepackage{multicol}
\usepackage{colortbl}
\usepackage{CJKutf8}

\DeclareMathOperator*{\argmax}{arg\,max}

\DeclareMathOperator{\logadd}{logadd}
\newcolumntype{?}{!{\vrule width 1pt}}

\aclfinalcopy 

\title{Effective Neural Solution for Multi-Criteria Word Segmentation}

\author{Han He$^1$, Lei Wu$^2$, Hua Yan$^3$  \\ 
\textbf{Zhimin Gao}$^4$, \textbf{Yi Feng}$^5$, \textbf{George Townsend}$^6$ \\
$^{1,2}$Computer and Software Engineering, Houston, Texas, U.S.A\\
$^4$Computer Science Department, Houston, Texas, U.S.A\\
$^{5,6}$Department of Computer Science, Alsoma University, Canada \\
{\small \tt $^{1,2,3}$\{heh1996,wul,yan\}@uhcl.edu}\\ {\small \tt $^4$zgao5@uh.edu, $^{5,6}$\{feng,townsend\}@algomau.ca}
}

\date{}

\begin{document}
\maketitle
\begin{abstract}
We present a simple yet elegant solution to train a single joint model on multi-criteria corpora for Chinese Word Segmentation (CWS). Our novel design requires no private layers in model architecture, instead, introduces two artificial tokens at the beginning and ending of input sentence to specify the required target criteria. The rest of the model including Long Short-Term Memory (LSTM) layer and Conditional Random Fields (CRFs) layer remains unchanged and is shared across all datasets, keeping the size of parameter collection minimal and constant. On Bakeoff 2005 and Bakeoff 2008 datasets, our innovative design has surpassed both single-criterion and multi-criteria state-of-the-art learning results. To the best knowledge, our design is the first one that has achieved the latest high performance on such large scale datasets. Source codes and corpora of this paper are available on GitHub\footnote{{\scriptsize \url{https://github.com/hankcs/multi-criteria-cws}}}.
\end{abstract}

\section{Introduction}

Unlike English language with space between every word, Chinese language has no explicit word delimiters. Therefore, Chinese Word Segmentation (CWS) is a preliminary pre-processing step for Chinese language processing tasks. Following \citet{Xue:2003ti}, most approaches consider this task as a sequence tagging task, and solve it with supervised learning models such as Maximum Entropy (ME) \cite{Jin2005A} and Conditional Random Fields (CRFs) \cite{Lafferty2001Conditional,peng2004chinese}. These early models require heavy handcrafted feature engineering within a fixed size window.

With the rapid development of deep learning, neural network word segmentation approach arose to reduce efforts in feature engineering \cite{Zheng:2013wj,Collobert:2011tk,Pei:2014vx,Chen:2015wa,Cai:2016tg,2017arXiv170407047C}. \citet{Zheng:2013wj} replaced raw character with its embedding as input, adapted the sliding-window based sequence labeling \cite{Collobert:2011tk}. \citet{Pei:2014vx} extended \citet{Zheng:2013wj}'s work by exploiting tag embedding and bigram features. \citet{Chen:2015wa} employed LSTM to capture long-distance preceding context. Noteworthily, a novel word-based approach \cite{Cai:2016tg,2017arXiv170407047C} was proposed to model candidate segmented results directly. Despite the outstanding runtime performance, their solution required the max word length $L$ to be a fixed hyper-parameter and replaced those words that longer than $L$ into a unique character. Thus their performance relies on an expurgation of long words, which is not practical.

Novel algorithms and deep models are not omnipotent. Large-scale corpus is also important for an accurate CWS system. Although there are many segmentation corpora, these datasets are annotated in different criteria, making it hard to fully exploit these corpora, which are shown in Table~\ref{different-criteria-table}.

\begin{table}[h]
\begin{center}
\begin{tabular}{|c|c|c|c|c|c|}
\hline 
\small{Corpora}&\small{Li}&\small{Le}&\small{reaches}&\small{Benz}&\small{Inc}\\
\hline 
\small{pku}&{ \begin{CJK}{UTF8}{gbsn}\small{李}\end{CJK}}&{ \begin{CJK}{UTF8}{gbsn}\small{乐}\end{CJK}}&{ \begin{CJK}{UTF8}{gbsn}\small{到达}\end{CJK}}&{ \begin{CJK}{UTF8}{gbsn}\small{奔驰}\end{CJK}}&{ \begin{CJK}{UTF8}{gbsn}\small{公司}\end{CJK}}\\
\hline \small{msr}&\multicolumn{2}{c|}{{ \begin{CJK}{UTF8}{gbsn}\small{李乐}\end{CJK}}}&{ \begin{CJK}{UTF8}{gbsn}\small{到达}\end{CJK}}&\multicolumn{2}{c|}{{ \begin{CJK}{UTF8}{gbsn}\small{奔驰公司}\end{CJK}}}\\
\hline
\small{as}&\multicolumn{2}{c|}{{ \begin{CJK}{UTF8}{bsmi}\small{李樂}\end{CJK}}}&{ \begin{CJK}{UTF8}{bsmi}\small{到達}\end{CJK}}&{ \begin{CJK}{UTF8}{bsmi}\small{賓士}\end{CJK}}&{ \begin{CJK}{UTF8}{bsmi}\small{公司}\end{CJK}}\\
\hline
\small{cityu}&\multicolumn{2}{c|}{{ \begin{CJK}{UTF8}{bsmi}\small{李樂}\end{CJK}}}&{ \begin{CJK}{UTF8}{bsmi}\small{到達}\end{CJK}}&{ \begin{CJK}{UTF8}{bsmi}\small{平治}\end{CJK}}&{ \begin{CJK}{UTF8}{bsmi}\small{公司}\end{CJK}}\\
\hline
\end{tabular}
\end{center}
\caption{\label{different-criteria-table} Illustration of different segmentation criteria of SIGHAN bakeoff 2005. }
\end{table}

Recently, \citet{2017arXiv170407556C} designed an adversarial multi-criteria learning framework for CWS. However, their models have several complex architectures, and are not comparable with the state-of-the-art results. 

In this paper, we propose a smoothly jointed multi-criteria learning solution for CWS by adding two artificial tokens at the beginning and ending of input sentence to specify the required target criteria. We have conducted various experiments on $8$ segmentation criteria corpora from SIGHAN Bakeoff 2005 and 2008. Our models improve performance by transferring learning on heterogeneous corpora. The final scores have surpassed previous multi-criteria learning, $2$ out of $4$ even have surpassed previous preprocessing-heavy state-of-the-art single-criterion learning results. 

The contributions of this paper could be summarized as: 
\begin{itemize}
\item Proposed an simple yet elegant solution to perform multi-criteria learning on multiple heterogeneous segmentation criteria corpora;
\item 2 out of 4 datasets have surpassed the state-of-the-art scores on Bakeoff 2005;
\item Extensive experiments on up to $8$ datasets have shown that our novel solution has significantly improved the performance.
\end{itemize}

\section{Related Work}

In this section, we review the previous works from 2 directions, which are Chinese Word Segmentation and multi-task learning.

\subsection{Chinese Word Segmentation}

Chinese Word Segmentation has been a well-studied problem for decades \cite{Huang2007Chinese}. After pioneer \citet{Xue:2003ti} transformed CWS into a character-based tagging problem, \citet{peng2004chinese} adopted CRF as the sequence labeling model and showed  its  effectiveness.  Following these pioneers, later sequence labeling based works \cite{Tseng2005A,Zhao:2006vi,Zhao2010A,sun2012fast} were proposed. Recent neural models \cite{Zheng:2013wj,Pei:2014vx,Chen:2015wa,Dong:2016bl,2017arXiv170407556C} also followed this sequence labeling fashion.

\subsection{Multi-Task Learning}

Compared to single-task learning, multi-task learning is relatively harder due to the divergence between tasks and heterogeneous annotation datasets. Recent works have started to explore joint learning on Chinese word segmentation or part-of-speech tagging. \citet{Jiang:2009ch} stacked two classifiers together. The later one used the former's prediction as additional features. \citet{Sun:2012up} proposed a structure-based stacking model in which one tagger was designed to refine another tagger's prediction.  These early models lacked a unified loss function and suffered from error propagation. 

\citet{Qiu:2013uc} proposed to learn a mapping function between heterogeneous corpora. \citet{Li:2015jo,Chao:2015dw} proposed and utilized coupled sequence labeling model which can directly learn and infer two heterogeneous annotations simultaneously. These works mainly focused on exploiting relationships between different tagging sets, but not shared features. 

\citet{2017arXiv170407556C} designed a complex framework involving sharing layers with Generative Adversarial Nets (GANs) to extract the criteria-invariant features and dataset related private layers to detect criteria-related features. This research work didn't show great advantage over previous state-of-the-art single-criterion learning scores.

Our solution is greatly motivated by Google's Multilingual Neural Machine Translation System, for which \citet{Johnson:2016va} proposed an extremely simple solution without any complex architectures or private layers. They added an artificial token corresponding to parallel corpora and train them jointly, which inspired our design.

\section{Neural Architectures for Chinese Word Segmentation}

A prevailing approach to Chinese Word Segmentation is casting it to character based sequence tagging problem  \cite{Xue:2003ti,sun2012fast}. One commonly used tagging set is $\mathcal{T} =  \{\text{B}, \text{M}, \text{E}, \text{S}\}$, representing the \textbf{b}egin, \textbf{m}iddle, \textbf{e}nd of a word, or single character forming a word. Given a sequence $\mathbf{X}$ with $n$ characters as $\mathbf{X} = (\mathbf{x}_1, \mathbf{x}_2, \ldots, \mathbf{x}_n)$, sequence tagging based CWS is to find the most possible tags $\mathbf{Y}^* = \{\mathbf{y}_1^*, \dots, \mathbf{y}_n^*\}$:
\begin{equation}
\mathbf{Y}^* = \argmax_{\mathbf{Y} \in \mathcal{T}^n} p (\mathbf{Y} | \mathbf{X}), \label{eq:cws-argmax}
\end{equation}

We model them jointly using a conditional random field, mostly following the architecture proposed by \citet{Lample:2016vz}, via stacking Long Short-Term Memory Networks (LSTMs) \cite{hochreiter1997long} with a CRFs layer on top of them.

We'll introduce our neural framework bottom-up. The bottom layer is a character Bi-LSTM (bidirectional Long Short-Term Memory Network) \cite{Graves:2005kt} taking character embeddings as input, outputs each character's contextual feature representation:
\begin{equation}
	\mathbf{h}_{t} =\text{Bi-LSTM}(\mathbf{X}, t)
\end{equation} 

After a contextual representation $\mathbf{h}_{t}$ is generated, it will be decoded to make a final segmentation decision. We employed a Conditional Random Fields (CRF) \cite{Lafferty2001Conditional} layer as the inference layer. 

First of all, a linear score function $s(\mathbf{X}, t) \in \mathbb{R}^{|\mathcal{T}|}$ is used to assign a local score for each tag on $t$-th character:
\begin{equation}
	s(\mathbf{X}, t) = \mathbf{W}_s^\top \mathbf{f}_t + \mathbf{b}_s
\end{equation}
where $\mathbf{f}_t=[\mathbf{h}_t; \mathbf{e}_t]$ is the concatenation of Bi-LSTM hidden state and bigram feature embedding $\mathbf{e}_t$, $\mathbf{W}_s \in \mathbb{R}^{d_f \times |\mathcal{T}|}$ and $\mathbf{b}_s \in \mathbb{R}^{|\mathcal{T}|}$ are trainable parameters. 

Then, for a sequence of predictions:
\begin{equation}
\mathbf{Y} = (y_1, y_2, \ldots, y_n)	
\end{equation}
first order linear chain CRFs employed a Markov chain to define its global score as:
\begin{equation}
	s(\mathbf{X}, \mathbf{Y})=\sum_{i=0}^{n} A_{y_i, y_{i+1}} + \sum_{i=1}^{n} P_{i, y_i}
\end{equation}
where $\mathbf{A}$ is a transition matrix such that $A_{i, j}$ represents the score of a transition from the tag $y_i$ to tag $y_j$. $y_0$ and $y_n$ are the \textit{start} and \textit{end} tags of a sentence, that are added to the tagset additionaly. $\mathbf{A}$ is therefore a square matrix of size $4+2$.

Finally, this global score is normalized to a probability in Equation (\ref{eq:cws-argmax}) via a softmax over all possible tag sequences:
\begin{equation}
	p(\mathbf{Y} | \mathbf{X}) = \frac{
e^{s(\mathbf{X}, \mathbf{Y})}
}{
\sum_{\mathbf{\widetilde{Y}} \in \mathbf{Y_X}} e^{s(\mathbf{X}, \mathbf{\widetilde{Y}})}
}
\end{equation}

In decoding phase, first order linear chain CRFs only model bigram interactions between output tags, so the maximum of a posteriori sequence $\mathbf{Y}^*$ in Eq.~\ref{eq:cws-argmax} can be computed using dynamic programming.

\section{Elegant Solution for Multi-Criteria Chinese Word Segmentation}

For closely related multiple task learning like multilingual translation system, \citet{Johnson:2016va} proposed a simple and practical solution. It only needs to add an artificial token at the beginning of the input sentence to specify the required target language, no need to design complex private encoder-decoder structures.

We follow their spirit and add two artificial tokens at the beginning and ending of input sentence to specify the required target criteria. For instance, sentences in SIGHAN Bakeoff 2005 will be designed to have the following form:

\begin{table}[h]
\begin{center}
\begin{tabular}{|c|c|}
\hline
\small{Corpora}&\small{Li Le reaches Benz Inc}\\
\hline
\small{PKU}&{ \begin{CJK}{UTF8}{gbsn}\scriptsize{$<$pku$>$ 李 乐 到达 奔驰 公司 $<$/pku$>$}\end{CJK}}\\
\hline
\small{MSR}&{ \begin{CJK}{UTF8}{gbsn}\scriptsize{$<$msr$>$ 李乐 到达 奔驰公司 $<$/msr$>$}\end{CJK}}\\
\hline
\small{AS}&{ \begin{CJK}{UTF8}{bsmi}\scriptsize{$<$as$>$ 李樂 到達 賓士 公司 $<$/as$>$}\end{CJK}}\\
\hline
\small{CityU}&{ \begin{CJK}{UTF8}{bsmi}\scriptsize{$<$cityu$>$ 李樂 到達 平治 公司 $<$/cityu$>$}\end{CJK}}\\
\hline
\end{tabular}
\end{center}
\caption{\label{criteria-table} Illustration of adding artificial tokens into 4 datasets on SIGHAN Bakeoff 2005. To be fair, these $<$dataset$>$ and $<$/dataset$>$ tokens will be removed when computing scores.}
\end{table}

 These artificial tokens specify which dataset the sentence comes from. They are treated as normal tokens, or more specifically, a normal character. With their help, instances from different datasets can be seamlessly put together and jointly trained, without extra efforts. These two special tokens are designed to carry criteria related information across long dependencies, affecting the context representation of every character, and finally to produce segmentation decisions matching target criteria. At test time, those tokens are used to specify the required segmentation criteria. Again, they won't be taken into account when computing performance scores.

\section{Training}

The training procedure is to maximize the log-probability of the gold tag sequence:
\begin{align}
\log(p(\mathbf{Y} | \mathbf{X})) 
&= \text{score}(\mathbf{X}, \mathbf{Y}) \notag\\
&- \underset{{\mathbf{\widetilde{Y}} \in \mathbf{Y_X}}}{\logadd}\ \text{score}(\mathbf{X}, \mathbf{\widetilde{Y}}), \label{eq:train-crf}
\end{align}
where $\mathbf{Y_X}$ represents all possible tag sequences for a sentence $\mathbf{X}$.

\section{Experiments}

We conducted various experiments to verify the following questions:
\begin{enumerate}
\item 	Is our multi-criteria solution capable of learning heterogeneous datasets?
\item 	Can our solution be applied to large-scale corpus groups consisting of tiny and informal texts?
\item 	More data, better performance?
\end{enumerate}

Our implementation is based on Dynet \cite{neubig2017dynet}, a dynamic neural net framework for deep learning. Additionally, we implement the CRF layer in Python, and integrated the official score script to verify our scores.

\subsection{Datasets}

To explore the first question, we have experimented on the 4 prevalent CWS datasets from SIGHAN2005 \cite{emerson_second_2005} as these datasets are commonly used by previous state-of-the-art research works. To challenge question 2 and 3, we applied our solution on SIGHAN2008 datasets \cite{moe2008fourth}, which are used to compare our approach with other state-of-the-art multi-criteria learning works under a larger scale. 

All datasets are preprocessed by replacing the continuous English characters and digits with a unique token. For training and development sets, lines are split into shorter sentences or clauses by punctuations, in order to make faster batch. 

Specially, the Traditional Chinese corpora CityU, AS and CKIP are converted to Simplified Chinese using the popular Chinese NLP tool HanLP\footnote{\url{https://github.com/hankcs/HanLP}}.

\subsection{Results on SIGHAN bakeoff 2005}

Our baseline model is Bi-LSTM-CRFs trained on each datasets separately. Then we improved it with multi-criteria learning. The final F$_1$ scores are shown in Table~\ref{bakeoff-result}. 

\begin{table} \setlength{\tabcolsep}{3pt}
\centering
\small
\begin{tabular}{c|cccc}
\hline
Models&PKU & MSR & CityU & AS\\
\hline
\citet{Tseng2005A} & 95.0 & 96.4 & - & - \\
\citet{zhang_chinese_2007} & 95.0 & 96.4 & - & - \\
\citet{zhao_unsupervised_2008}&95.4 & \textbf{97.6} & 96.1 & \textbf{95.7}\\
\citet{sun2009a}& 95.2&97.3&-&-\\
\citet{sun2012fast}& 95.4&97.4&-&-\\
\citet{zhang_exploring_2013}$^{\clubsuit}$        &96.1&97.4&-&-\\
\citet{Chen:2015wj}$^{\spadesuit}$ &94.5&95.4&-&-\\
\citet{Chen:2015wa}$^{\spadesuit}$ &94.8&95.6&-&-\\
\citet{2017arXiv170407556C}                    &94.3&96.0&-&94.8\\
\citet{2017arXiv170407047C}$^{\diamondsuit}$&95.8  & 97.1 & 95.6 & 95.3\\
\hline
baseline &95.2 & 97.3 & 95.1 & 94.9\\
+multi&\textbf{95.9}  & 97.4 & \textbf{96.2} & 95.4\\
\hline
\end{tabular}
\caption{
Comparison with previous state-of-the-art models of results on all four Bakeoff-2005 datasets. Results with $\clubsuit$ used external dictionary or corpus, with $\spadesuit$ are from \citet{Cai:2016tg}'s runs on their released implementations without dictionary, with ${\diamondsuit}$ expurgated long words in test set.
}
\label{bakeoff-result}
\end{table}

According to this table, we find that multi-criteria learning boosts performance on every single dataset. Compared to single-criterion learning models (baseline), multi-criteria learning model (+multi) outperforms all of them by up to $1.1\%$. Our joint model doesn't rob performance from one dataset to pay another, but share knowledge across datasets and improve performance on all datasets.

\subsection{Results on SIGHAN bakeoff 2008}

SIGHAN bakeoff 2008 \cite{moe2008fourth} provided as many as $5$ heterogeneous corpora. With another $3$ non-repetitive corpora from SIGHAN bakeoff 2005, they form a large-scale standard dataset for multi-criteria CWS benchmark. We repeated our experiments on these $8$ corpora and compared our results with state-of-the-art scores, as listed in Table~\ref{sighan08}.

\begin{table*}\small
\centering
\begin{tabular}{|c|*{9}{c|}>{\columncolor[gray]{.8}}c|}
\hline
  \multicolumn{2}{|c|}{Models}&
	MSR    &AS     &PKU    &CTB    &CKIP   &CITYU  &NCC    &SXU    &Avg.\\
\hline
\hline

\multicolumn{11}{|l|}{Single-Criterion Learning} \\
\hline

\multirow{4}*{ \citet{2017arXiv170407556C}} 
&P  &95.70  &93.64  &93.67  &95.19  &92.44  &94.00  &91.86  &95.11  &93.95  \\
&R  &95.99  &94.77  &92.93  &95.42  &93.69  &94.15  &92.47  &95.23  &94.33  \\
&F  &95.84  &94.20  &93.30  &95.30  &93.06  &94.07  &92.17  &95.17  &94.14  \\
 \hline

\multirow{4}*{Ours} 
&P  &97.17  &95.28  &94.78  &95.14  &94.55  &94.86  &93.43  &95.75      &95.12  \\
&R  &97.40  &94.53  &95.66  &95.28  &93.76  &94.16  &93.74  &95.80      &95.04  \\
&F  &97.29  &94.90  &95.22  &95.21  &94.15  &94.51  &93.58  &95.78      &95.08  \\
 \hline
 \hline
\multicolumn{11}{|l|}{Multi-Criteria Learning} \\
\hline
\multirow{4}*{ \citet{2017arXiv170407556C}} 
&P  &95.95  &94.17  &94.86  &96.02  &93.82  &95.39  &92.46  &96.07  &94.84  \\
&R  &96.14  &95.11  &93.78  &96.33  &94.70  &95.70  &93.19  &96.01  &95.12  \\
&F  &96.04  &94.64  &94.32  &\textbf{96.18}  &94.26  &95.55  &92.83  &96.04  &94.98  \\
\hline
\multirow{4}*{Ours} 
&P  &97.38  &96.01  &95.37  &95.69  &96.21  &95.78  &94.26  &96.54   &95.82  \\
&R  &97.32  &94.94  &96.19  &96.00  &95.27  &95.43  &94.42  &96.44   &95.64  \\
&F  &\textbf{97.35}  &\textbf{95.47}  &\textbf{95.78}  &95.84  &\textbf{95.73}  &\textbf{95.60}  &\textbf{94.34}  &\textbf{96.49}   &\textbf{95.73}  \\
\hline
\end{tabular}
\caption{Results on test sets of 8 standard CWS datasets.
Here, P, R, F indicate the precision, recall, $\text{F}_1$ value respectively. The maximum $\text{F}_1$ values are highlighted for each dataset.
}\label{sighan08}
\end{table*}

In the first block for single-criterion learning, we can see that our implementation is generally more effective than \citet{2017arXiv170407556C}'s. In the second block for multi-criteria learning, this disparity becomes even significant. And we further verified that every dataset benefit from our joint-learning solution. We also find that more data, even annotated with different standards or from different domains, brings better performance. Almost every dataset benefits from the larger scale of data. In comparison with large datasets, tiny datasets gain more performance growth.

\section{Conclusions and Future Works}

\subsection{Conclusions}

In this paper, we have presented a practical way to train multi-criteria CWS model. This simple and elegant solution only needs adding two artificial tokens at the beginning and ending of input sentence to specify the required target criterion. All the rest of model architectures, hyper-parameters, parameters and feature space are shared across all datasets. Experiments showed that our multi-criteria model can transfer knowledge between differently annotated corpora from heterogeneous domains. Our system is highly end-to-end, capable of learning large-scale datasets, and outperforms the latest state-of-the-art multi-criteria CWS works.

\subsection{Future Works}

Our effective and elegant multi-criteria learning solution can be applied to sequence labeling tasks such as POS tagging and NER. We plan to conduct more experiments of using our effective technique in various application domains.

\bibliographystyle{acl_natbib}
\bibliography{references}

\end{document}